\documentclass{article}
\usepackage{spconf,amsmath,graphicx}

\usepackage{enumitem}
\setlist{nosep, leftmargin=14pt}

\usepackage{color}
\usepackage[table]{xcolor}
\usepackage{amsmath}
\usepackage{amssymb}
\usepackage{multirow}
\usepackage{color}
\usepackage{bm}
\usepackage{bbm}
\usepackage{arydshln} 
\usepackage{mathtools}
\usepackage{booktabs} 
\usepackage{pifont}

\title{Continuous 3D Spatial Transcriptomics Reconstruction via Histology-Informed Implicit Neural Representations}
\name{
\shortstack{
Kaito Shiku$^{1,2}$ \quad
Ryoma Bise$^{1}$ \quad
Ahtisham Fazeel Abbasi$^{2,3}$ \quad
Yuichiro Iwashita$^{2,3,4}$ \\
Shumpei Takezaki$^{1,2}$ \quad
Andreas Dengel$^{2,3,4,5}$ \quad
Muhammad Nabeel Asim$^{2,4,5}$
}
}

\address{
$^{1}$ Kyushu University \quad
$^{2}$ DFKI GmbH \quad
$^{3}$ RPTU University Kaiserslautern-Landau \\
$^{4}$ Osaka Metropolitan University \quad
$^{5}$ IntelligentX GmbH
}

\begin{document}
\ninept
\maketitle
\begin{abstract}
Three-dimensional spatial transcriptomics (3D ST) enables the analysis of gene-expression patterns across tissue depths. 
However, 3D ST measurements are acquired from discrete 2D tissue sections, which are often sparsely sampled due to high acquisition costs. 
To address this limitation, we propose a histology-informed multi-task implicit neural representation (HIM-INR) for continuous 3D gene-expression reconstruction from sparse ST observations. HIM-INR jointly reconstructs gene-expression profiles and histological representations using an INR with a shared backbone, leveraging densely sampled histological images as auxiliary spatial supervision. Once trained, HIM-INR predicts gene expression at arbitrary 3D coordinates without requiring histological images. Experiments on two 3D ST datasets demonstrate improved reconstruction over existing methods, particularly as the number of missing ST sections increases.
\end{abstract}


\section{Introduction}
\label{sec:intro}

Three-dimensional spatial transcriptomics (3D ST) \cite{mo2024tumour} has emerged as a promising approach for investigating the spatial organization of gene expression within tissues. 
Building on conventional 2D ST~\cite{staahl2016visualization},
which measures gene expression in individual tissue sections,
3D ST enables the characterization of molecular interactions
across tissue depths and the spatial extent and connectivity
of pathological regions. Such capabilities offer new opportunities to investigate cell-to-cell communication across sections, characterize tumor invasion and inflammatory infiltration, and reveal spatial relationships between tissue architecture and molecular states that cannot be fully captured by individual 2D sections~\cite{mo2024tumour, pentimalli2025combining}.

Despite the potential of 3D ST, gene expression is measured
only on discrete tissue sections. Furthermore, the high cost
of ST measurements restricts the number of sections that can
be profiled, resulting in sparse observations along the tissue
depth axis. Previous studies have aligned ST measurements
from serial tissue sections into a common spatial coordinate
system to reconstruct tissue structures along the depth
axis~\cite{zeira2022alignment, zhou2023integrating}.
However, gene-expression information remains restricted
to the observed sections, leaving the continuous distribution
of gene expression between them unresolved.

\begin{figure*}[t]
\centering
\includegraphics[width=0.85\linewidth]{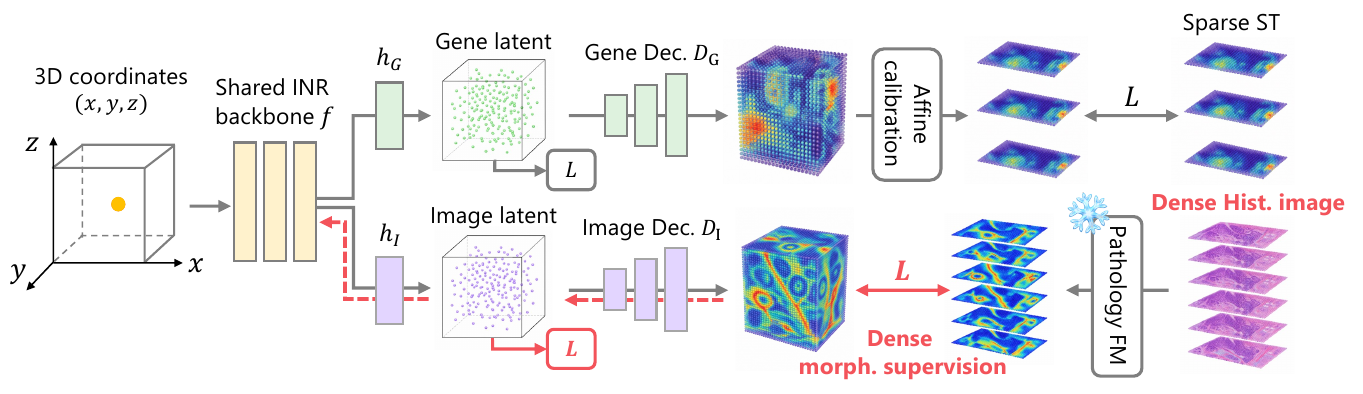}
\vspace{-4mm}
\caption{Overview of the proposed histology-informed multi-task INR (HIM-INR). Gene-expression profiles and histology image features are compressed into modality-specific latent spaces. A shared INR backbone with modality-specific heads and decoders reconstructs both modalities from 3D coordinates, enabling dense image-only sections to provide auxiliary spatial supervision between sparse ST observations.}
\label{fig:method}
\vspace{-1mm}
\end{figure*}

These limitations motivate the reconstruction of a continuous
3D gene-expression field from sparsely and discretely observed ST sections.
However, this task remains challenging.
While gene expression can vary substantially with changes in tissue structure across sections, no direct information on gene expression is available between sparsely sampled ST sections.
Consequently, not only conventional neighborhood-based interpolation
methods but also deep learning-based approaches~\cite{li2024high, zhang2026sinter3d}
that model nonlinear relationships between spatial positions
and gene expression face difficulties in accurately reconstructing
gene expression in unobserved regions.

To address this challenge, our key idea is to leverage
histological images, which provide detailed morphological
information about tissue structure.
Gene-expression patterns are known to be associated with tissue structure and morphology~\cite{he2020integrating, chung2024accurate}. 
Moreover, histological images are commonly acquired alongside ST measurements and are often available from more densely sampled tissue sections due to their substantially lower acquisition cost~\cite{schroeder2025scaling}. 
Thus, histological information can provide valuable cues
about variations in tissue morphology, even at locations
where ST measurements are unavailable, potentially facilitating
continuous 3D gene-expression reconstruction.

In this study, we propose a histology-informed multi-task implicit neural representation
(HIM-INR) framework for reconstructing gene-expression profiles
at arbitrary 3D coordinates from sparsely sampled ST sections,
as illustrated in Fig.~\ref{fig:method}.
We base our method on an INR model, which represents continuous
signals by learning a function that maps spatial coordinates
to corresponding signal values~\cite{sitzmann2020implicit}.
Using the INR as a shared backbone, our framework jointly reconstructs
gene-expression profiles and histological representations extracted
using a pretrained pathology foundation model designed to capture
tissue morphology~\cite{chen2024towards}.
By incorporating histological reconstruction as an auxiliary task,
the shared INR backbone receives additional spatial supervision
from densely sampled histological sections, even at locations
where ST measurements are unavailable.
This encourages the model to capture continuous variations in
tissue morphology and learn shared spatial representations
for gene-expression and histological reconstruction.
Importantly, inference requires only 3D coordinates, enabling
gene-expression reconstruction even at locations where neither
ST measurements nor histological images are available.

We evaluate the proposed framework on two publicly available 3D ST datasets and compare it with conventional interpolation methods and INR-based reconstruction approaches.
Experimental results demonstrate that our method outperforms existing approaches under most evaluation settings, highlighting the effectiveness of incorporating histological information for reconstructing gene expression profiles at unobserved sections. 
In particular, the benefits of histological supervision become more pronounced as the number of missing ST sections increases, highlighting its potential for reconstructing continuous gene expression fields from sparse observations.

\section{Related Work}
\noindent
{\bf 3D ST Reconstruction.}
Recent studies have explored 3D ST analysis by integrating transcriptomic measurements from serial tissue sections. 
Most approaches focus on aligning consecutive sections to reconstruct their spatial organization~\cite{zeira2022alignment, zhou2023integrating}. 
Although these methods enable the analysis of spatial relationships across observed sections, they do not reconstruct gene-expression patterns in unobserved regions between sections, limiting continuous 3D analysis.

Recent methods have addressed gene-expression reconstruction
between sparsely observed tissue sections.
STAGE~\cite{li2024high} aligns the latent space of a
gene-expression autoencoder with spatial coordinates,
enabling prediction at arbitrary locations by subsequently
feeding only spatial coordinates into the decoder.
SINTER3D~\cite{zhang2026sinter3d} uses an
INR~\cite{sitzmann2020implicit} to predict gene expression
at unobserved 3D coordinates.
However, learning spatial continuity solely from
transcriptomic measurements remains challenging when
ST sections are sparsely sampled along the depth axis.
In contrast, we explore the potential of leveraging
more densely sampled histological images as auxiliary
supervision to encourage the INR to capture morphological
variations between observed ST sections.

\noindent
{\bf INR for ST Analysis.}
INR parameterizes continuous signals using neural networks that map spatial coordinates to signal values~\cite{sitzmann2020implicit}.
Recent studies have applied INR to impute missing gene-expression values and reconstruct expression at unobserved locations in 2D ST~\cite{luo2025stinr,zhu2025suica}.
STINR~\cite{luo2025stinr} directly regresses gene-expression profiles from spatial coordinates, whereas SUICA~\cite{zhu2025suica} addresses high-dimensional expression reconstruction by learning a low-dimensional latent representation predicted using INR.
However, these methods primarily focus on reconstruction
within individual 2D tissue sections rather than continuous
modeling of gene expression in 3D space across aligned sections.

\noindent
{\bf Histology-based ST Estimation.}
Predicting gene expression from histological images has
emerged as a major research direction in medical image
analysis~\cite{he2020integrating, chung2024accurate}.
These studies have demonstrated a strong relationship
between tissue morphology and gene expression through
image-based prediction.
However, their reliance on histological images at inference
time prevents prediction at locations where images are
unavailable, limiting their applicability to our task.

\begin{table*}[t]
\centering
\renewcommand{\arraystretch}{1.0}

\caption{Quantitative comparison of gene-expression reconstruction at withheld sections. PCC is computed on the top 50 and 250 HVGs and averaged over three random seeds. Best and second-best results are shown in bold and underlined, respectively.}

\label{tab:comparison}

\setlength{\tabcolsep}{6.5pt}

\begin{tabular}{l cc cc cc}
\toprule
\multirow{2}{*}{\textbf{Method}}
& \multicolumn{2}{c}{\textbf{3DST-RA}~\cite{vickovic2022three}}
& \multicolumn{2}{c}{\textbf{HemiBrainA}~\cite{ortiz2020molecular}}
& \multicolumn{2}{c}{\textbf{HemiBrainB}~\cite{ortiz2020molecular}} \\

\cmidrule(lr){2-3}
\cmidrule(lr){4-5}
\cmidrule(lr){6-7}

& HVGs50 & HVGs250
& HVGs50 & HVGs250
& HVGs50 & HVGs250 \\

\midrule

NN
& 0.1779 & 0.1385
& 0.0966 & 0.1240
& 0.0958 & 0.0741 \\

KNN
& 0.2609 & 0.2148
& 0.1449 & 0.1921
& 0.1288 & 0.1114 \\

STAGE~\cite{li2024high}
& 0.2743 & 0.2041
& \textbf{0.2344} & \underline{0.3095}
& 0.1398 & \underline{0.1525} \\

3D-STINR~\cite{luo2025stinr}
& 0.2202 & 0.2425
& 0.1395 & 0.2113
& \underline{0.1703} & 0.1453 \\

3D-SUICA~\cite{zhu2025suica}
& \underline{0.3409} & \underline{0.3044}
& 0.1883 & 0.2664
& 0.1518 & 0.1513 \\

SINTER3D~\cite{zhang2026sinter3d}
& 0.1847 & 0.2288
& 0.1724 & 0.2326
& 0.1627 & 0.1399 \\

\rowcolor{gray!15}
\textbf{Ours}
& \textbf{0.3711} & \textbf{0.3145}
& \underline{0.2253} & \textbf{0.3101}
& \textbf{0.1824} & \textbf{0.1685} \\

\bottomrule
\end{tabular}

\end{table*}

\section{Histology-Informed Multi-task INR}
\label{sec:method}

\subsection{Overview}
\label{sec:overview}

We propose HIM-INR to reconstruct gene expression at arbitrary 3D coordinates using sparsely observed ST sections and more densely sampled histological images.
As illustrated in Fig.~\ref{fig:method}, HIM-INR consists of
two reconstruction branches for gene expression and histology
image features, sharing a SIREN-based INR backbone~\cite{sitzmann2020implicit}.
Inspired by Zhu et al.~\cite{zhu2025suica}, we address the difficulty of modeling high-dimensional vectors with INR by learning spatial continuity in low-dimensional latent spaces for both modalities.
The predicted latent representations are then decoded into
gene-expression profiles and histology image features.
During training, the model is jointly optimized using reconstruction losses for both modalities.
Importantly, the histological-feature reconstruction branch is also trained on sections without ST measurements, encouraging the shared INR backbone to capture tissue structural information that guides gene-expression reconstruction in unobserved regions.

We consider sparsely observed ST measurements,
denoted by
$\mathcal{S}_G=\{(\mathbf{p}_i,\mathbf{g}_i)\}_{i\in\Omega_G}$,
and more densely sampled histological observations,
denoted by
$\mathcal{S}_I=\{(\mathbf{p}_i,\mathbf{v}_i)\}_{i\in\Omega_I}$.
Here, $\Omega_G$ and $\Omega_I$ denote the index sets
of observed ST and histology image features, respectively,
with $\Omega_G\subseteq\Omega_I$.
The spatial coordinate $\mathbf{p}_i\in\mathbb{R}^3$
specifies the location of each observation,
while $\mathbf{g}_i\in\mathbb{R}^G$ and
$\mathbf{v}_i\in\mathbb{R}^D$ represent gene-expression
profiles and image-derived features, respectively,
with $G$ genes and $D$ feature dimensions.
To encourage the INR to capture tissue morphology rather
than low-level pixel information, we reconstruct features
extracted by a pretrained pathology foundation model~\cite{chen2024towards}
instead of raw histological images.
All serial tissue sections are registered into a common
3D coordinate system prior to training.
HIM-INR reconstructs gene-expression profiles and
histology image features at arbitrary 3D coordinates:
\begin{equation}
(\hat{\mathbf{g}}_i,\hat{\mathbf{v}}_i)
=
\mathrm{HIM\text{-}INR}(\mathbf{p}_i).
\label{eq:himinr}
\end{equation}

\subsection{Latent Representations of ST and Histology Features}
To enable the INR to learn spatial continuity in low-dimensional latent spaces rather than directly modeling high-dimensional observations, we employ autoencoders to obtain latent representations of both gene-expression profiles and histology image features.
We independently pretrain MLP-based autoencoders for each modality using a mean squared error (MSE) reconstruction loss, with latent dimensions $d_G$ and $d_I$, respectively.
After pretraining, the encoders are frozen and used to extract latent representations
$\mathbf{z}_i^G\in\mathbb{R}^{d_G}$ and
$\mathbf{z}_i^I\in\mathbb{R}^{d_I}$
from the corresponding observations.
These representations serve as fixed targets for subsequent INR training.

\subsection{Histology-informed Multi-task Learning}

We jointly learn continuous representations of sparsely observed ST and densely sampled histological images through a shared INR backbone $f$.
By incorporating image-only sections during training, the shared backbone captures spatial variations in tissue structure even at locations without ST observations, thereby guiding gene-expression reconstruction.

Given a 3D coordinate $\mathbf{p}_i$,
the shared backbone extracts a spatial representation, which is mapped to modality-specific latent spaces by two prediction heads, $h_G$ and $h_I$:
\begin{equation}
\hat{\mathbf{z}}_i^G = h_G(f(\mathbf{p}_i)), \quad
\hat{\mathbf{z}}_i^I = h_I(f(\mathbf{p}_i)).
\end{equation}
The predicted latent representations are subsequently decoded into gene-expression profiles and histology image features using modality-specific decoders $D_G$ and $D_I$:
\begin{equation}
\hat{\mathbf{g}}_i = D_G(\hat{\mathbf{z}}_i^G), \quad
\hat{\mathbf{v}}_i = D_I(\hat{\mathbf{z}}_i^I).
\end{equation}

To mitigate the impact of section-specific technical variations
and batch effects on learning spatial continuity,
we introduce slice- and gene-specific affine calibration.
For an ST observation $i$ belonging to section $s(i)$,
the calibrated gene-expression profile is computed as
$\tilde{\mathbf{g}}_i =
\mathbf{a}_{s(i)}\odot\hat{\mathbf{g}}_i +
\mathbf{b}_{s(i)}$,
where $\mathbf{a}_{s(i)},\mathbf{b}_{s(i)}\in\mathbb{R}^G$
are learnable scale and bias vectors, respectively, and
$\odot$ denotes element-wise multiplication.

The entire model is trained by minimizing reconstruction
losses in both latent and output spaces for each modality,
defined as follows:
\begin{equation}
\begin{aligned}
\mathcal{L}_G
&=
\frac{1}{|\Omega_G|}
\sum_{i\in\Omega_G}
\left[
\|\hat{\mathbf{z}}_i^G-\mathbf{z}_i^G\|_2^2
+
\|\tilde{\mathbf{g}}_i-\mathbf{g}_i\|_2^2
\right],\\
\mathcal{L}_I
&=
\frac{1}{|\Omega_I|}
\sum_{i\in\Omega_I}
\left[
\|\hat{\mathbf{z}}_i^I-\mathbf{z}_i^I\|_2^2
+
\|\hat{\mathbf{v}}_i-\mathbf{v}_i\|_2^2
\right].
\end{aligned}
\end{equation}
The overall objective is
\begin{equation}
\mathcal{L}_{\mathrm{total}}
=
\mathcal{L}_G+\lambda\mathcal{L}_I,
\end{equation}
where $\lambda$ balances the contributions of the two modalities.
For image-only locations, only the latent-space and
output-space image reconstruction losses are applied.
By updating the shared INR backbone, these losses
encourage the learning of spatial continuity even
in regions without ST observations.

During inference, we directly use the gene-expression
predictions $\hat{\mathbf{g}}_i$ without applying
the slice-specific affine calibration, as it is
introduced during training to mitigate the effects
of technical variations and batch effects
on learning spatial continuity.

\section{Experiments}
\label{sec:experiments}
\noindent
{\bf Datasets and Experimental Settings.}
Although histological images are commonly acquired alongside
ST measurements~\cite{domanskyi2024nextflow},
large-scale, publicly available 3D ST datasets with paired
histological images remain limited. We therefore select two datasets with relatively large numbers of sections, \textbf{3DST-RA}~\cite{vickovic2022three} and \textbf{HemiBrain}~\cite{ortiz2020molecular}. To evaluate our method, we artificially withhold tissue sections to simulate missing observations.
 
3DST-RA contains paired ST measurements and histological images from serial synovial tissue sections of patients with rheumatoid arthritis.
We select the patient with the largest number of sections, consisting of seven consecutive sections numbered from 1 to 7. ST measurements are retained for Sections 1, 4, and 7. Among the remaining four sections, three are used as image-only observations, while one is completely withheld for evaluation. We repeat this evaluation four times, each time selecting a different section to be completely withheld.

The HemiBrain dataset contains serial brain sections from three mice. We use two samples, HemiBrainA and HemiBrainB, containing 42 and 9 sections, respectively. We retain both ST measurements and histological images at every fourth section. Among the three intermediate sections, the first and third are used as image-only observations, while the second is withheld for evaluation.

Prior to training and evaluation, adjacent tissue sections are aligned using XFeat-based feature matching~\cite{potje2024xfeat} and RANSAC-based similarity transformations~\cite{fischler1981random}. These transformations are sequentially composed to register all sections to a common coordinate system, with the middle section as the reference. Evaluation-section histological images are also used for registration to ensure accurate spatial correspondence between predicted and ground-truth gene-expression profiles, but are excluded from model training, inference, and performance evaluation.

\begin{table*}[t]
    \centering
    \renewcommand{\arraystretch}{1.0}
    \caption{Ablation study on histological supervision (HIM).
    PCC is evaluated on image-unavailable locations using the top 50 and 250 HVGs.
    }
    \label{tab:ablation}

    \scalebox{1}{
    \setlength{\tabcolsep}{6.5pt}
    \begin{tabular}{l c cc cc cc}
    \toprule
    \multirow{2}{*}{\textbf{Method}}
    & \multirow{2}{*}{\textbf{HIM}}
    & \multicolumn{2}{c}{\textbf{3DST-RA}~\cite{vickovic2022three}}
    & \multicolumn{2}{c}{\textbf{HemiBrainA}~\cite{ortiz2020molecular}}
    & \multicolumn{2}{c}{\textbf{HemiBrainB}~\cite{ortiz2020molecular}} \\

    \cmidrule(lr){3-4}
    \cmidrule(lr){5-6}
    \cmidrule(lr){7-8}

    & & HVGs50 & HVGs250
    & HVGs50 & HVGs250
    & HVGs50 & HVGs250 \\

    \midrule

    Ours w/o HIM
    & \ding{55}
    & 0.3593 & 0.3054
    & \textbf{0.2259} & 0.3052
    & 0.1742 & 0.1645 \\

    \rowcolor{gray!15}
    \textbf{Ours}
    & \ding{51}
    & \textbf{0.3711} & \textbf{0.3145}
    & 0.2253 & \textbf{0.3101}
    & \textbf{0.1824} & \textbf{0.1685} \\

    \bottomrule
    \end{tabular}
    }
    \vspace{-1mm}
\end{table*}

\begin{figure}[t]
\centering
\includegraphics[width=0.73\linewidth]{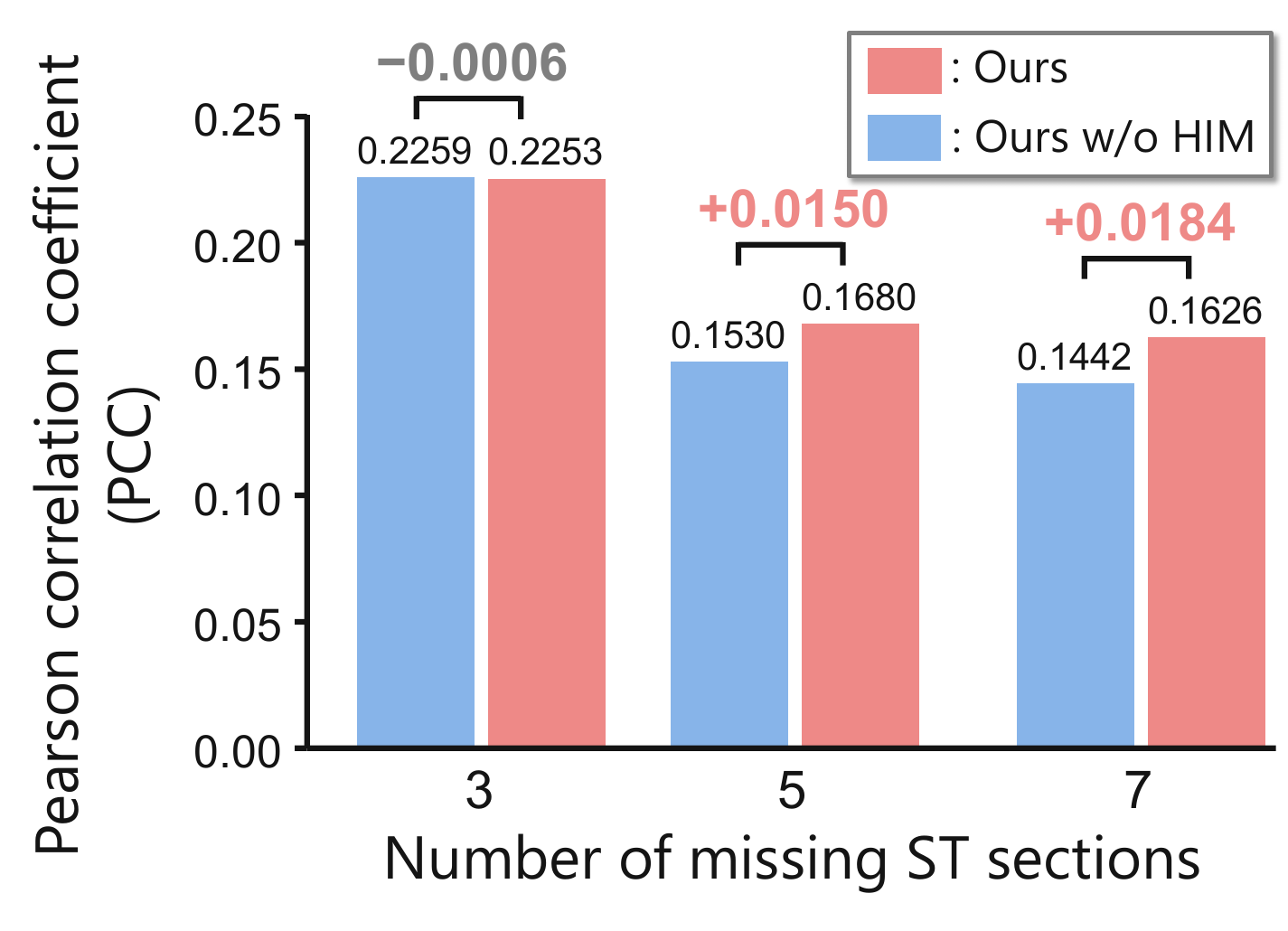}
\vspace{-3mm}
\caption{Effect of ST sparsity on HemiBrainA (top 50 HVGs).
The number of missing ST sections between observed sections
is varied from 3 to 7.
Values above the bars indicate the PCC differences
between Ours and Ours w/o HIM (Ours $-$ Ours w/o HIM).
}
\label{fig:ablation}
\vspace{-1mm}
\end{figure}

\noindent
{\bf Implementation Details and Evaluation Metrics.}
All experiments used PyTorch~\cite{Paszke2019PyTorchAI} with Adam~\cite{adam}. Histology image features were extracted using the frozen UNI2-h~\cite{chen2024towards}. The best model was selected based on validation performance, using 5\% of ST spots randomly held out from each observed section. Hyperparameters were set to a learning rate of $10^{-3}$, up to 1,000 epochs, a batch size of 512, latent dimensions $d_G=d_I=5$, and a loss weight $\lambda=1$.

To evaluate reconstruction performance while minimizing slice-specific batch effects, we compute the Pearson correlation coefficient (PCC)~\cite{chung2024accurate} between predicted and ground-truth expression across spatial locations for each gene within each section, then average across genes and sections. Following standard ST prediction protocols~\cite{jaume2024hest}, we evaluate the top 50 and 250 highly variable genes (HVGs) to mitigate measurement noise. We report mean performance over three random seeds.

\noindent
{\bf Comparisons.}
We compare our method with six baselines.
1) ``NN'' and 2) ``KNN'' interpolate gene expression from the nearest and $k=5$ nearest observed spots, respectively.
3) ``STAGE''~\cite{li2024high} aligns autoencoder latent representations with spatial coordinates, enabling prediction at unobserved locations through its decoder.
4) ``3D-STINR''~\cite{luo2025stinr} and 5) ``3D-SUICA''~\cite{zhu2025suica} are INR-based methods originally developed for 2D ST and extended to 3D using spatially aligned coordinates across sections.
3D-STINR directly regresses gene expression from 3D coordinates, whereas 3D-SUICA first learns gene-expression embeddings using slice-wise graph autoencoders and then trains an INR to predict these embeddings from 3D coordinates, which are subsequently decoded into gene expression.
6) ``SINTER3D''~\cite{zhang2026sinter3d} predicts gene expression using an INR with additional intermediate-feature reconstruction losses to encourage spatial continuity.

Table~\ref{tab:comparison} presents the quantitative comparison.
Simple interpolation methods, NN and KNN, show poor performance.
STAGE performs well on HemiBrainA but shows limited performance on other datasets, indicating a lack of robustness across datasets. This may be because explicitly aligning gene-expression latent representations with spatial coordinates restricts their ability to capture complex expression patterns.
Similarly, 3D-STINR and SINTER3D show limited performance, potentially due to the difficulty of directly regressing high-dimensional expression vectors using INR.
In contrast, 3D-SUICA achieves relatively consistent performance, suggesting that low-dimensional latent modeling facilitates INR training.
Our method achieves the best performance on all metrics for 3DST-RA and HemiBrainB and on HVGs250 for HemiBrainA, demonstrating the effectiveness of histological supervision for continuous 3D reconstruction from sparse ST observations across datasets.

\begin{table}[t]
    \centering
    \renewcommand{\arraystretch}{1.0}
    \caption{Comparison with oracle image-based prediction (UNI2-h+LP), which uses true histological images at the evaluation sections. PCC is evaluated on the top 250 HVGs. 
    }
    \label{tab:image_comparison}
    \setlength{\tabcolsep}{6pt}

    \scalebox{0.8}{
    \begin{tabular}{l c c c}
    \toprule
    \textbf{Method}
    & \textbf{3DST-RA}~\cite{vickovic2022three}
    & \textbf{HemiBrainA}~\cite{ortiz2020molecular}
    & \textbf{HemiBrainB}~\cite{ortiz2020molecular} \\

    \midrule

     UNI2-h+LP
    & 0.2726
    & 0.2744
    & 0.1646 \\

    \rowcolor{gray!15}
    \textbf{Ours}
    & \textbf{0.3145}
    & \textbf{0.3101}
    & \textbf{0.1685} \\

    \bottomrule
    \end{tabular}
    }
\end{table}

\noindent
{\bf Ablation Study.}
To evaluate the effect of morphological supervision,
we compare our method with ``Ours w/o HIM'',
which excludes image-based supervision.
Both reconstruct gene expression in the same
low-dimensional latent space, differing only in
whether the shared INR backbone receives additional
supervision from histological images.
As shown in Table~\ref{tab:ablation}, our method achieves relatively consistent and competitive performance across datasets, suggesting that dense histological supervision helps the INR learn spatial continuity more reliably than gene expression alone.

To further assess histological supervision, we vary the number of missing ST sections (3, 5, and 7) on HemiBrainA, the largest dataset. The middle missing section is held out for evaluation, while the others provide only histological images. As shown in Fig.~\ref{fig:ablation}, both methods perform comparably with three missing sections, whereas our method increasingly outperforms Ours w/o HIM as the number of missing sections increases. This suggests that histological supervision is particularly beneficial when sparse ST observations make learning spatial continuity from gene expression alone challenging.

\noindent\textbf{Validity of Histology as Multi-task Supervision.}
To validate the use of densely sampled histological images as auxiliary supervision in multi-task learning, we consider an alternative that first interpolates histology image features at unobserved sections and then uses them as direct inputs to predict gene expression.
To evaluate this alternative without errors from histological feature interpolation, we train a linear probe on frozen UNI2-h features from ST-observed sections (UNI2-h+LP) and evaluate it using ground-truth histological images at the evaluation sections. Importantly, these images are unavailable in the actual reconstruction setting and are never provided to our method.
As shown in Table~\ref{tab:image_comparison}, our method outperforms the
UNI2-h+LP baseline even when ground-truth histological
images are provided at the evaluation sections.
This suggests that histological supervision can be
beneficial for learning continuous spatial representations
beyond this simple image-based prediction strategy.

\section{Conclusion}
We proposed HIM-INR, a histology-informed multi-task INR for continuous 3D gene-expression reconstruction from sparse ST observations. By jointly learning gene-expression and histological representations through a shared INR backbone, our method leverages densely sampled histological images as auxiliary supervision to capture spatial continuity between observed ST sections. Experiments on two 3D ST datasets demonstrated improved reconstruction over existing methods, with increasing benefits as ST observations become sparser. Importantly, our method enables gene-expression prediction at arbitrary 3D coordinates without requiring histological images at inference time.

\section{Compliance with Ethical Standards}
This study involved secondary analysis of publicly available human
and mouse spatial transcriptomics datasets (3DST-RA and HemiBrain).
The original studies reported approval from the relevant ethics
committees, and written informed consent was obtained from the
human participants~\cite{vickovic2022three, ortiz2020molecular}.
No new human or animal experiments were conducted in this study.

\section{Acknowledgments}
\label{sec:acknowledgments}
This work was supported by  JST BOOST, Japan Grant Number JPMJBS2406, ASPIRE Grant Number JPMJAP2403.
A. Dengel and M. N. Asim are affiliated with IntelligentX GmbH. IntelligentX GmbH did not provide financial support for this work. The authors have no other relevant financial or non-financial interests to disclose.
Anthropic Claude Code assisted with code development,
and OpenAI ChatGPT assisted with language polishing
and revision throughout the manuscript.
The authors take full responsibility for the code
and the final manuscript.
\bibliographystyle{IEEEbib}
\bibliography{refs}

\end{document}